\documentclass{ecai}
\usepackage{times}
\usepackage{graphicx}
\usepackage{latexsym}

\usepackage{diagbox}
\usepackage{multirow}
\usepackage{tikz}
\usepackage{tikz-cd}
\usetikzlibrary{matrix,calc,arrows,arrows.meta}
\usepackage{arydshln}
\usepackage{booktabs}
\usepackage{url}
\usepackage{color}

\begin{document}

\title{Applying the Closed World Assumption to \SUMO{}-based FOL Ontologies for Effective Commonsense Reasoning}


\author{Javier \'{A}lvez\institute{LoRea Group, University of the Basque Country (UPV/EHU), email: javier.alvez@ehu.eus} \and Itziar Gonzalez-Dios\institute{Ixa Group, HiTZ Center, University of the Basque Country (UPV/EHU), email: itziar.gonzalezd@ehu.eus} \and German Rigau\institute{Ixa Group, HiTZ Center, University of the Basque Country (UPV/EHU), email: \mbox{german.rigau@ehu.eus}} }



\newcommand{\WORDNET}{WordNet}
\newcommand{\SUMO}{SUMO}
\newcommand{\DOLCE}{DOLCE}
\newcommand{\CYC}{Cyc}
\newcommand{\TPTPSUMO}{TPTP-SUMO}
\newcommand{\ADIMENSUMO}{Adimen-SUMO}


\newcommand{\quartertab}{\hspace{5pt}}
\newcommand{\minitab}{\quartertab\quartertab}
\newcommand{\tab}{\minitab\minitab}
\newcommand{\textConstant}[1]{{\it{#1}}}
\newcommand{\textVariable}[1]{{\it{?#1}}}
\newcommand{\textFunction}[1]{{\it{#1}}}
\newcommand{\textPredicate}[1]{{\it{#1}}}
\newcommand{\connective}[1]{\bf #1 \;}
\newcommand{\predicate}[1]{\rm #1}
\newcommand{\constant}[1]{\rm #1}
\newcommand{\variable}[1]{\tt ?#1}
\newcommand{\true}{\it true}
\newcommand{\false}{\it false}


\newcommand{\SUMOObjectSymbol}{o}
\newcommand{\SUMOClassSymbol}{c}
\newcommand{\SUMOIndividualRelationSymbol}{r}
\newcommand{\SUMOIndividualAttributeSymbol}{a}
\newcommand{\SUMOClassOfAttributesSymbol}{A}

\newcommand{\SUMOObject}[1]{\textConstant{#1}$_\SUMOObjectSymbol$}
\newcommand{\SUMOClass}[1]{\textConstant{#1}}
\newcommand{\SUMORelation}[1]{\textConstant{#1}$_\SUMOIndividualRelationSymbol$}
\newcommand{\SUMOAttribute}[1]{\textConstant{#1}$_\SUMOIndividualAttributeSymbol$}
\newcommand{\SUMOClassOfAttributes}[1]{\textConstant{#1}$_\SUMOClassOfAttributesSymbol$}

\newcommand{\allDirectSubclassesOf}[1]{all\_direct\_subclasses\_of(#1)}


\newcommand{\synset}[3]{{\textConstant{#1}}}
\newcommand{\WORDNETRelation}[1]{{\it{#1}}}


\newcommand{\synsetTikZ}[3]{{\it{#1}}}

\newcommand{\SUMOObjectTikZ}[1]{{\constant{#1}_\SUMOObjectSymbol}}
\newcommand{\SUMOClassTikZ}[1]{{\textConstant{#1}}}
\newcommand{\SUMOIndividualRelationTikZ}[1]{{\constant{\it #1}_\SUMOIndividualRelationSymbol}}
\newcommand{\SUMOClassOfRelationsTikZ}[1]{{\constant{\it #1}_\SUMOClassofRelationsSymbol}}
\newcommand{\SUMOIndividualAttributeTikZ}[1]{{\constant{#1}_\SUMOIndividualAttributeSymbol}}
\newcommand{\SUMOClassOfAttributesTikZ}[1]{{\constant{#1}_\SUMOClassOfAttributesSymbol}}

\newcommand{\equivalenceMappingTikZOfConcept}[1]{{#1}\hspace{-4pt}\equivalenceMappingSymbol}
\newcommand{\subsumptionMappingTikZOfConcept}[1]{{#1}\subsumptionMappingSymbol}
\newcommand{\instanceMappingTikZOfConcept}[1]{{#1}\instanceMappingSymbol}
\newcommand{\negatedEquivalenceMappingTikZOfConcept}[1]{{#1}\negatedEquivalenceMappingSymbol}
\newcommand{\negatedSubsumptionMappingTikZOfConcept}[1]{{#1}\negatedSubsumptionMappingSymbol}


\newcommand{\equivalenceMappingSymbol}{=}
\newcommand{\subsumptionMappingSymbol}{+}
\newcommand{\instanceMappingSymbol}{@}
\newcommand{\negatedMappingSymbol}{\widehat{~}}
\newcommand{\negatedEquivalenceMappingSymbol}{\widehat{\equivalenceMappingSymbol}}
\newcommand{\negatedSubsumptionMappingSymbol}{\widehat{\subsumptionMappingSymbol}}

\newcommand{\equivalenceMapping}[1]{{\textConstant{#1}$\equivalenceMappingSymbol$}}
\newcommand{\subsumptionMapping}[1]{{\textConstant{#1}$\subsumptionMappingSymbol$}}
\newcommand{\instanceMapping}[1]{{\textConstant{#1}$\instanceMappingSymbol$}}
\newcommand{\negatedEquivalenceMapping}[1]{{\textConstant{#1}$\negatedEquivalenceMappingSymbol$}}
\newcommand{\negatedSubsumptionMapping}[1]{{\textConstant{#1}$\negatedSubsumptionMappingSymbol$}}

\newcommand{\updated}[1]{{\color{blue}{#1}}}

\newcommand{\question}[1]{{\color{green}{#1}}}

\newcommand{\correct}[1]{{\color{red}{#1}}}

\maketitle
\bibliographystyle{ecai}

\begin{abstract}
Most commonly, the {\it Open World Assumption} is adopted as a standard strategy for the design, construction and use of ontologies. This strategy limits the inferencing capabilities of any system because non-asserted statements ({\it missing knowledge}) could be assumed to be alternatively true or false. As we will demonstrate, this is especially the case of first-order logic (FOL) ontologies where non-asserted statements is nowadays one of the main obstacles to its practical application in automated commonsense reasoning tasks. In this paper, we investigate the application of the {\it Closed World Assumption} (CWA) to enable a better exploitation of FOL ontologies by using state-of-the-art automated theorem provers. To that end, we explore different CWA formulations for the structural knowledge encoded in a FOL translation of the \SUMO{} ontology, discovering that almost $30$~\% of the structural knowledge is missing. We evaluate these formulations on a practical experimentation using a very large commonsense benchmark obtained from \WORDNET{} through its mapping to \SUMO{}. The results show that the competency of the ontology improves more than $50$~\% when reasoning under the CWA. Thus, applying the CWA automatically to FOL ontologies reduces their ambiguity and more commonsense questions can be answered.
\end{abstract}

\section{Introduction}

\noindent Large knowledge-bases and complex ontologies are being used in a wide range of knowledge based systems \cite{AnH08} that require practical commonsense reasoning \cite{Cha99,noy2001ontology,StS09,Aquin11,vossen2016newsreader}. To represent this knowledge, the most prominent and fundamental logical formalism is the {\it first-order predicate calculus}, or first-order logic (FOL) for short. The semantics of FOL, and thus also of {\it Description Logics} (DL), operates under the {\it Open World Assumption} (OWA) allowing monotonic reasoning \cite{drummond2006open}. OWA considers that statements which are {\it not} logical consequences of a given knowledge base are not necessarily considered false but possible. Therefore, statements that are false or impossible must be clearly stated as so in the ontology. The OWA presumes incomplete knowledge about the domain being modelled. 
Thus, ontologies basically encode positive information about the modelled world since the number of negative facts vastly exceeds the number of positive ones. In fact, under the OWA, it is totally unfeasible to explicitly represent all such negative information in the ontology.

Otherwise, the {\it Closed World Assumption} (CWA) presumes perfect knowledge about the domain being modelled. CWA is a common non-monotonic technique that allows to deal with negative information in knowledge bases and data bases \cite{Rei78}. In fact, commonsense reasoning is non-monotonic: the addition of new knowledge can invalidate conclusions drawn before the addition \cite{Grimm2007knowledge}. 

There is a considerable computational and representational advantage to reason under the CWA since negative information should be inferred by default \cite{reiter1980logic}. 
The Careful CWA (CCWA) is an extension of the CWA \cite{gelfond1986negation}. It allows us to restrict the effects of closing the world by specifying the predicates which may be affected by the CWA rule in indefinite databases. 


Nowadays OWL 2 \cite{W3C12} is currently one of the most common formal knowledge representation formalism,  but it is unable to fully cope with general upper ontologies like \CYC{} \cite{Matuszek+'06}, \DOLCE{} \cite{Gangemi+'02} or \SUMO{} \cite{Niles+Pease'01} since full FOL expressivity or higher is required. Further, CWA cannot be entirely applied to DL ontologies, but some approximations have been proposed in the literature \cite{EIL08,MoR08,KSH11}.


In order to provide advanced reasoning support to large FOL conversions of expressive ontologies \cite{RRG05,HoV06,PeS07,ALR12}
state-of-the-art automated theorem provers (ATP) for FOL such as Vampire \cite{KoV13} or E \cite{Sch02} have proven its efficiency by implementing many sophisticated techniques like axiom selection \cite{HoV11}. However, the semi-decidability of FOL and the poor scalability of the known decision procedures have been usually identified as the main drawbacks for the practical use of FOL ontologies. 



In this paper, we report on our empirical research applying the Careful CWA, that was originally conceived for indefinite databases, to the structural knowledge of a FOL ontology. In particular, we propose two complementary strategies for the application of the CCWA to the structural knowledge about classes represented in a FOL conversion of the top levels of \SUMO{} \cite{Niles+Pease'01}. To the best of our knowledge, this is the first attempt of applying the CWA to FOL ontologies since up to now the research on \SUMO{} has been developed under the OWA.

\begin{figure}[ht]
\centering
\begin{tikzpicture}[>=triangle 60]
\matrix[matrix of math nodes,column sep={-3pt},row sep={50pt,between origins},nodes={asymmetrical rectangle}] (s)
{
|[name=beverage]| \langle \synsetTikZ{beverage}{1}{n} \rangle & : & |[name=beverageMappingClass]| [ \SUMOClassTikZ{Beverage} ] \\
|[name=drinkingWater]| \langle \synsetTikZ{drinking\_water}{1}{n} \rangle & : & |[name=drinkingWaterMappingClass]| [ \SUMOClassTikZ{Water} ] \\
};
\draw[-latex] (drinkingWater) -- node[left] {\(\langle hyponym \rangle\)} (beverage);
\draw[-stealth,dotted] (drinkingWaterMappingClass) -- node[right] {\(?\)} (beverageMappingClass);
\end{tikzpicture}\caption{An example of competency question for \SUMO{} obtained from \WORDNET{}}
\label{fig:introduction}
\end{figure}
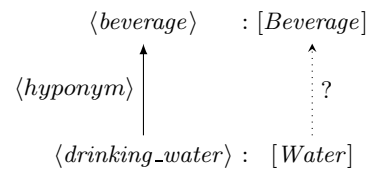

We test the original and the resulting versions of \SUMO{} by using the knowledge in \WORDNET{} \cite{Fellbaum'98} as gold standard. For this purpose, we build a benchmark by automatically deriving a very large set of {\it competency questions} (CQs) from \WORDNET{}  and its mapping to \SUMO{} \cite{Niles+Pease'03} on the basis of 7 manually created {\it question patterns (QPs)} \cite{ALR17}. These QPs focus on the main structural relations of \WORDNET{}, which are {\it hyponymy} and {\it antonymy}. The results show that applying carefully the CWA to the structural knowledge about classes in \SUMO{} improves the competency of the ontology more than 50~\% when reasoning on the same commonsense benchmark.
%
%

For instance, in Figure \ref{fig:introduction} we describe the CQ ``{\it Drinking water is a beverage}'' that results from the hyponymy pair of \WORDNET{} synsets \synset{drinking\_water}{1}{n} (the {\it hyponym}) and \synset{beverage}{1}{n} (the {\it hyperonym}), which are respectively connected to the \SUMO{} concepts \SUMOClass{Water} and \SUMOClass{Beverage}. From this CQ, we obtain two conjectures: the first one states that some
instances of \SUMOClass{Water} can 
be
instance of \SUMOClass{Beverage}, and the second one is its negation.
None of these conjectures are solved using \SUMO{}, but one of them is entailed depending on the strategy for the application of the CWA to \SUMO{}, concretely the CCWA to {\it subclass} and {\it disjoint} predicates. From now on, we will refer to CCWA as CWA.   

Our research empirically demonstrates the existence of large knowledge gaps in \SUMO{} and that the {\it missing knowledge} in \SUMO{}-based FOL ontologies is nowadays one of the main obstacles for its practical application in automated commonsense reasoning tasks. Anyway, our proposal is not intended to provide that missing knowledge but, nevertheless, it could help ontologists to complete the knowledge in the ontology.

%

%

The contributions of this paper are fourfold. First, we propose an effective method to apply automatically the CWA to the structural knowledge of \SUMO{}, which enables a really compact formalization.  Second, we perform a detailed analysis of the empirical results obtained when comparing the resulting versions of \SUMO{} with the original one on a very large commonsense benchmark with more than 14,000 CQs. This analysis demonstrates that the competency of the ontology can improve more than $50$~\% when reasoning under the CWA. Third, we provide a quantitative analysis of the structural knowledge about classes that still needs to be encoded in the top levels of \SUMO{}. Fourth, we discuss some suitable design criteria that enable the automated application of the CWA in FOL ontologies. 


{\it Outline of the paper}.  In Section \ref{section:sota} we present \SUMO{} and its translations into FOL; in Section \ref{section:Completion}, we describe our approaches for the application of the CWA to {\it subclass} and {\it disjoint}; in Section \ref{section:experimentation} we report on the experimental results that we discuss in Section \ref{section:discussion}; and we conclude in Section \ref{section:Conclusions} by outlining the future work.

\section{ \SUMO{} and its FOL versions} \label{section:sota}


\SUMO{}\footnote{\url{http://www.ontologyportal.org}} \cite{Niles+Pease'01} is a well-known upper level ontology proposed as a starter  document by the IEEE Standard Upper Ontology Working Group. \SUMO{} is expressed in SUO-KIF (Standard Upper Ontology Knowledge Interchange Format \cite{Pea09}), which is a dialect of KIF (Knowledge Interchange Format \cite{Richard+'92}). The syntax of both KIF and SUO-KIF goes beyond FOL and, therefore, \SUMO{} axioms cannot be directly used by FOL ATPs without a suitable transformation. 

To the best of our knowledge, there are two main proposals for the translation of the two upper levels of \SUMO{} into FOL formulas that are described in \cite{PeS07,pease2010large} and \cite{ALR12} respectively. Both proposals have been developed under the OWA and are currently included in the {\it Thousands of Problems for Theorem Provers} (TPTP) problem library\footnote{\url{http://www.tptp.org}} \cite{Sut09}.


The knowledge of \SUMO{}, and therefore of its translations into FOL, is organized around the notions of {\it classes} and {\it particulars}. The main structural knowledge about classes is provided by the predicates i) {\it subclass}, which is defined as a partial order relation (reflexive, transitive and anti-symmetric), and ii) {\it disjoint}, which is defined as symmetric and irreflexive. The predicate {\it subclass} provides the classical concept of relation inclusion between classes, while the predicate {\it disjoint} relates incompatible classes: in \ADIMENSUMO{}, incompatible classes cannot share any common instance or subclass. In \SUMO{}, particulars are introduced by the predicate {\it instance}.

From the axiomatization of \SUMO{}, particulars are inherited by superclasses (inheritance of {\it instance} via {\it subclass}). Additionally,  every pair of disjoint classes do not share any instance and are not subclass of each other. Further, \SUMO{} includes some additional predicates that provide structural knowledge about disjoint classes: specifically, {\it partition} and {\it disjointDecomposition}.

In the experiments, we will use \ADIMENSUMO{} the FOL version of \SUMO{} that has proved to be more competent
\cite{AGR18,black19,Siebert19}. Currently, \ADIMENSUMO{} consists of 8,291 formulas, out of them 5,255 are atomic, and defines $2,169$ classes.

Since \ADIMENSUMO{} has been developed under the OWA, sometimes negative knowledge is not inferable. For example, although considering both the explicit and implicit knowledge in \ADIMENSUMO{}, it is not possible to infer whether \SUMOClass{SentientAgent} (``{\it An agent that has rights but may or may not have responsibilities and the ability to reason}'') and \SUMOClass{Sandwich} (``{\it Any food which consists of two or more pieces of bread and some sort of filling between the two pieces of bread}'') are related by {\it subclass}/{\it disjoint} or not. From now on, we say that pairs of classes are {\it non-asserted pairs} (or {\it missing knowledge}) if \ADIMENSUMO{} cannot entail whether they are related or not by {\it subclass}/{\it disjoint}. We have obtained an upper-bound of the amount of missing structural knowledge in \ADIMENSUMO{}. For this purpose, we have considered both the explicit and implicit knowledge in \ADIMENSUMO{} as follows: first, we have used {\it ad hoc} tools by focusing on the structural knowledge from \ADIMENSUMO{}; second, we have used ATPs with the whole knowledge of \ADIMENSUMO{}. Among the total of $4,704,561$ ($2,169^2$) different pairs of classes, it is possible to infer that a) $18,374$ ($0.39$~\%) pairs of classes are related by {\it subclass} (thus, not related by {\it disjoint}), b) $3,304,246$ ($70.23$~\%) pairs of classes are related by {\it disjoint} (thus, not related by {\it subclass}) and c) $62,069$ ($1.32$~\%) pairs of classes are not related by {\it disjoint} (because those pairs of classes share some instance/subclass). Consequently, there are at most $1,381,941$\footnote{$1,381,941 = 4,704,561 - (18,374 + 3,304,246)$} non-asserted {\it subclass} pairs ($29.37$~\%) and $1,338,246$\footnote{$1,338,246 = 4,704,561 - (3,304,246 + 62,069$)} non-asserted {\it disjoint} pairs ($28.45$~\%). In other words, in the worst case scenario almost $30$~\% of the structural knowledge about classes is missing in \ADIMENSUMO{}.





\section{Completing \ADIMENSUMO{}} \label{section:Completion}

In this section, we describe different applications of the CWA to the structural knowledge about classes in \ADIMENSUMO{} in order to reduce missing knowledge. We also provide the amount of new formulas that are required in each case. Concretely, we focus on the predicates {\it subclass} and {\it disjoint}. In the case of {\it subclass}, we apply a single strategy in which every non-asserted {\it subclass} pairs in \ADIMENSUMO{} are assumed not to be related by {\it subclass} (see Subsection \ref{subsection:CompletingSubclass}). 
 With respect to {\it disjoint}, we apply two complementary strategies by assuming disjointness/non-disjointness to non-asserted pairs (see Subsections \ref{subsection:CompletingDisjoint} and \ref{subsection:CompletingNonDisjoint} respectively).

\subsection{Applying the CWA to \textbf{\textit{subclass}}} \label{subsection:CompletingSubclass}

In this subsection, we describe our proposal for the application of the CWA to {\it subclass} by assuming that the non-asserted {\it subclass} pairs are not related by {\it subclass}.

The application of the CWA to {\it subclass} is based on the set of class pairs that are explicitly related by {\it subclass}: {\it direct} subclasses. From now on, we denote by $\allDirectSubclassesOf{c}$ the set of all the \ADIMENSUMO{} classes that are explicitly defined to be direct subclasses of an \ADIMENSUMO{} class $c$.

In order to apply the CWA to {\it subclass}, we conveniently adapt the {\it data base completion} method proposed in \cite{Cla78}.  For this adaptation, we implicitly adopt the {\it Domain Closure Asssumption} (DCA) (that is, closed domain of classes) and assume that the domain of classes is finite: the domain only includes the classes that are explicitly introduced by the ontology.

Our adaptation of the data base completion relies on the fact that {\it subclass} is defined as a partial order relation in \SUMO{}. This implies that given two classes $c$ and $c'$ such that $c'$ is direct subclass of $c$ (that is, $c' \in \allDirectSubclassesOf{c}$), every subclass of $c'$ is also subclass of $c$ (by transitivity) and $c$ is subclass of itself (by reflexivity):
\begin{eqnarray}
 & \forall ?x \forall ?y \forall ?z \; ( \; subclass(?x,?y) \wedge subclass(?y,?z) ) \to \label{eq:transitivity} \\
 & \hspace{100pt} subclass(?x,?z) \; ) \nonumber \\
 & \forall ?x \; ( \; subclass(?x,?x) \; ) \label{eq:reflexivity}
\end{eqnarray}
Further, we also know that any superclass of $c$ (except of $c$ itself) is not subclass of $c$ (by antisymmetry):
\begin{equation}
\forall ?x \forall ?y \; ( \; ( subclass(?x,?y) \wedge subclass(?y,?x) ) \to ?x = ?y \; ) \label{eq:antisymmetry}
\end{equation}
Consequently, the {\it completed set of subclasses of an \ADIMENSUMO{} class $c$} can be defined as
\begin{equation}
\forall ?x \; ( \; subclass(?x,c) \leftrightarrow ( ?x = c \vee \bigvee_{i=1}^n subclass(?x,c_i) ) \; ) \label{eq:CompletionOfSubclass}
\end{equation}
where $\allDirectSubclassesOf{c} = \{ c_1, \ldots, c_n \}$. The reverse implication is already given by the ontology. In particular, by axioms (\ref{eq:transitivity}-\ref{eq:antisymmetry}). Thus, we only have to augment \ADIMENSUMO{} by including the direct implication.
For example, \SUMOClass{BloodCell} have two direct subclasses in \ADIMENSUMO{}, which are \SUMOClass{RedBloodCell} and \SUMOClass{WhiteBloodCell}:
\begin{eqnarray}
 & & subclass(RedBloodCell,BloodCell) \label{eq:RedBloodCell} \\
 & & subclass(WhiteBloodCell,BloodCell) \label{eq:WhiteBloodCell}
\end{eqnarray}
Therefore, we have that $\allDirectSubclassesOf{\SUMOClass{BloodCell}} = \{ \SUMOClass{RedBloodCell}, \SUMOClass{WhiteBloodCell} \}$ and the formula that results from (\ref{eq:CompletionOfSubclass}) to complete the information about the subclasses of \SUMOClass{BloodCell} in \ADIMENSUMO{} is:
\begin{eqnarray}
 & & \forall ?x \; ( \; subclass(?x,BloodCell) \to \label{eq:CompletionOfBloodCell} \\
 & & \hspace{70pt}( \; ?x = BloodCell \; \vee \nonumber \\
 & & \hspace{77pt}subclass(?x,RedBloodCell) \; \vee \nonumber \\
 & & \hspace{77pt}subclass(?x,WhiteBloodCell) \; ) \; ) \nonumber
\end{eqnarray}
It is worth noting that, although the transitive closure of a binary relation cannot ---in general--- be expressed in first-order logic, all the formulas involved in the proposed application of the DCA to \ADIMENSUMO{} classes (axioms (\ref{eq:transitivity}-\ref{eq:antisymmetry}) and (\ref{eq:RedBloodCell}-\ref{eq:CompletionOfBloodCell}) in the case of \SUMOClass{BloodCell}) are pure FOL formulas. 

In total, since \ADIMENSUMO{} inherits $2,169$ classes from \SUMO{} we have automatically augmented \ADIMENSUMO{} by including $2,169$ new formulas as the one above (one per \SUMO{} class), where we have used $4,705$ subclass atoms.


In this approach every non-asserted {\it subclass} pairs in \ADIMENSUMO{} are assumed not to be related by {\it subclass}. It is worth noting that the complementary strategy ---that is, assuming that every non-asserted {\it subclass} pairs in \ADIMENSUMO{} are related by {\it subclass}--- turns most of the classes into equal by antisymmetry.


%
%
%

\subsection{Applying the CWA to \textbf{\textit{disjoint}} by assuming disjointness} \label{subsection:CompletingDisjoint}

In this subsection, we describe our proposal for the application of the CWA to {\it disjoint} by assuming that the non-asserted {\it disjoint} pairs of classes are disjoint.



\begin{figure}
\centering
\begin{tikzpicture}[sibling distance=30pt]
  \node {\SUMOClass{Beverage}}
    child { node {\SUMOClass{Coffee}} }
    child { node {\SUMOClass{Milk}} }
    child { node {\SUMOClass{Tea}} }
    child { node {$\ldots$} };
\end{tikzpicture}
\begin{tikzpicture}[sibling distance=40pt]
  \node {\SUMOClass{CompoundSubstance}}
    child { node {\SUMOClass{Water}} }
    child { node {\SUMOClass{ChemicalSalt}} }
    child { node {$\ldots$} };
\end{tikzpicture}
\caption{Non-asserted {\it disjoint} subclasses of \SUMOClass{Beverage} and \SUMOClass{CompoundSubstance}}
\label{fig:BeverageCompoundSubstance}
\end{figure}
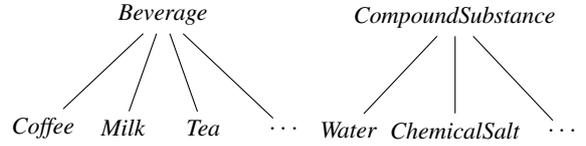
Formally, the application of the CWA by assuming disjointness can be described as follows: for any pair of non-asserted {\it disjoint} classes $c_1$ and $c_2$, we augment \ADIMENSUMO{} by stating that $c_1$ and $c_2$ are disjoint. For example, in Figure \ref{fig:BeverageCompoundSubstance} we show some of the subclasses of \SUMOClass{Beverage} and \SUMOClass{CompoundSubstance}, where  all the depicted subclasses of \SUMOClass{Beverage} are non-asserted {\it disjoint} with all the depicted subclasses of \SUMOClass{CompoundSubstance}. Hence, the above described application of the CWA to {\it disjoint} by assuming disjointness introduces, among others, the following formulas in \ADIMENSUMO{}:
\begin{eqnarray} 
 & & disjoint(Beverage,CompoundSubstance) \label{formula:CWADisjointBeverageCompoundSubstance} \\
\ & & disjoint(Beverage,Water) \label{formula:CWADisjointBeverageWater}
\end{eqnarray}
It is obvious that the second conjecture obtained from ``{\it Drinking water is a beverage}'' is entailed by the augmented version of \ADIMENSUMO{}, because \SUMOClass{Beverage} and \SUMOClass{Water} do not have any common instance/subclass if they are disjoint (see formula (\ref{formula:CWADisjointBeverageWater})).

\begin{table*}[ht!]
\caption{\label{table:summary} Summary of Experimentation Results}
\centering
\begin{tabular}{l|rrr|rrr|rrr}
\toprule
\multicolumn{1}{c|}{{\bf Competency}} & \multicolumn{3}{c|}{\bf OWA} & \multicolumn{3}{c|}{\bf CWA-D} & \multicolumn{3}{c}{\bf CWA-n-D} \\
\multicolumn{1}{c|}{{\bf Questions}} & \multicolumn{1}{c}{{\bf \#}} & \multicolumn{1}{c}{{\bf \%}} & \multicolumn{1}{c|}{{\bf T}} & \multicolumn{1}{c}{{\bf \#}} & \multicolumn{1}{c}{{\bf \%}} & \multicolumn{1}{c|}{{\bf T}} & \multicolumn{1}{c}{{\bf \#}} & \multicolumn{1}{c}{{\bf \%}} & \multicolumn{1}{c}{{\bf T}} \\
\midrule
noun \#1 (7,402) & 4,016 & 54.26~\% & 47.86 s. & 5,250 & 70.93~\% & 54.71 s. & ~~5,064 & 68.41~\% & 51.28 s. \\
noun \#2 (1,866) & 1,209 & 64.79~\% & 12.87 s. & 1,475 & 79.05~\% & 31.20 s. & 1,288 & 69.02~\% & 19.09 s. \\
verb \#1 (1,740) & 637 & 36.61~\%  & 72.79 s.& 1,454 & 83.56~\% & 41.94 s. & 1,340 & 77.01~\% & 51.42 s. \\
verb \#2 (299) & 144 & 48.16~\%  & 30.92 s.& 252 & 84.28~\% & 61.44 s. & 155 & 51.84~\% & 23.42 s. \\
antonym \#1 (65) & 36 & 55.38~\%  & 44.90 s.& 29 & 44.62~\% & 33.99 s. & 27 & 41.54~\% & 39.49 s. \\
antonym \#2 (504) & 152 & 30.16~\%  & 108.15 s.& 159 & 31.55~\% & 80.53 s. & 111 & 22.02~\% & 110.85 s. \\
antonym \#3 (2,448) & 1,091 & 44.57~\%  & 155.79 s.& 1,162 & 47.47~\% & 99.50 s. & 1,513 & 61.81~\% & 124.16 s. \\
\midrule
{\bf Total (14,324)} & {\bf 7,285} & {\bf 50.86~\%} & {\bf 61.30 s.} & {\bf 9,781} & {\bf 68.28~\%} & {\bf 55.12 s.} & {\bf 9,498} & {\bf 66.31~\%} & {\bf 58.75 s.} \\
\bottomrule
\end{tabular}
\end{table*}

In practice, most of the formulas that results from the application of the above method are redundant because of the axiomatization of {\it disjoint} in \SUMO{} and can be easily omitted. More specifically, two classes are related by {\it disjoint} iff there is no common instance or subclass. Hence, given a pair of disjoint classes $c_1$ and $c_2$, every subclass $c'_1$ of $c_1$ (resp. $c'_2$ of $c_2$) is disjoint with $c_2$ (resp. $c_1$) by the inheritance of {\it instance} via {\it subclass} and the transitivity of {\it subclass} since all the instances/subclasses of $c'_1$ (resp. $c'_2$) are also instance/subclass of $c_1$ (resp. $c_2$). Therefore, disjointness is inherited by subclasses. For example, the classes \SUMOClass{Beverage} and \SUMOClass{Water} are still inferred to be disjoint although augmenting \ADIMENSUMO{} by only formula (\ref{formula:CWADisjointBeverageCompoundSubstance}) (and not formula (\ref{formula:CWADisjointBeverageWater})), because \SUMOClass{Water} is defined as subclass of \SUMOClass{CompoundSubstance} in \ADIMENSUMO{}.

This way, we have augmented \ADIMENSUMO{} by adding $20,896$ non-redundant atomic formulas.

\subsection{Applying the CWA to \textbf{\textit{disjoint}} by assuming non-disjointness} \label{subsection:CompletingNonDisjoint}

Conversely, next we describe the application of the CWA to {\it disjoint} by assuming that the non-asserted {\it disjoint} pairs of classes are non-disjoint.

For this purpose, we  proceed similar to our previous strategy: for any pair of non-asserted {\it disjoint} classes $c_1$ and $c_2$, augment \ADIMENSUMO{} by stating that $c_1$ and $c_2$ are non-disjoint. Coming back to the example about the non-asserted {\it disjoint} classes in Figure \ref{fig:BeverageCompoundSubstance}, the above application of the CWA to {\it disjoint} by assuming non-disjointness would introduce, among others, the next formulas:
\begin{eqnarray} 
 & & \neg disjoint(Beverage,CompoundSubstance) \label{formula:CWANonDisjointExplicitBeverageCompoundSubstance} \\
 & & \neg disjoint(Beverage,Water) \label{formula:CWANonDisjointExplicitBeverageWater} \\
 & & \neg disjoint(Coffee,Water) \label{formula:CWANonDisjointExplicitCoffeeWater}
\end{eqnarray}
This time, the first conjecture that results from ``{\it Drinking water is a beverage}'' is entailed by the augmented version of \ADIMENSUMO{} due to formula (\ref{formula:CWANonDisjointExplicitBeverageWater}), since pairs of non-disjoint subclasses necessarily have some common instance/subclass. However, this way we would obtain, as before, many redundant formulas: given a pair of non-disjoint classes $c_1$ and $c_2$, it can be inferred in \ADIMENSUMO{} that all the superclasses $c'_1$ of $c_1$ (resp. $c'_2$ of $c_2$) are non-disjoint with $c_2$ (resp. $c_1$) by the inheritance of {\it instance} via {\it subclass} and the transitivity of {\it subclass}. That is, non-disjointness is inherited {\it upwards}. However, it is not easy to omit the redundant formulas in this case: for this purpose, we have to introduce non-disjoint pairs only between classes that do not have subclasses. This generates a very high number of new formulas. In addition, we need to check whether the classes are defined as disjoint. To verify this, two strategies can be followed: 1) stating for each pair if it is disjoint, for which it is necessary to use a lot of memory; 2) checking the hierarchy of classes. In this last case, there are two other options:
\begin{itemize}
\item By following a top-down strategy: given a branch, it is necessary to check all other non-disjoint ones, with many repetitions.
\item By following a bottom-up strategy: for each pair that is non-disjoint, it is necessary to check the whole branch from the leaves to the root, which is computationally expensive. 
\end{itemize}

In order to minimize the number of atomic formulas that are required to apply the CWA to {\it disjoint} by assuming non-disjointness, we introduce a new predicate ---{\it nonDisjoint}--- that states the {\it downwards} inheritance of non-disjointness. This new predicate is axiomatized as follows:
\begin{equation} \label{formula:nonDisjoint}
\begin{array}{l}
\forall ?x_1 \forall ?x_2 \forall ?y_1 \forall ?y_2 \; ( \; ( \; nonDisjoint(?x_1,?x_2) \; \wedge \\
\hspace{96pt} subclass(?y_1,?x_1) \; \wedge \\
\hspace{96pt} subclass(?y_2,?x_2) \; ) \; \to \\
\hspace{117pt} \neg disjoint(?y_1,?y_2) \; )  \\
\end{array}
\end{equation}

By means of this new predicate, we can proceed as follows: for any pair of 
classes $c_1$ and $c_2$ such that 
\begin{enumerate}
\item[{\it a})] the pair $c_1$ and $c_2$ is non-asserted {\it disjoint}, 
\item[{\it b})] any subclasses $c'_1$ and $c'_2$ of $c_1$ and $c_2$ respectively are either non-asserted {\it disjoint} or not related by {\it disjoint}
\end{enumerate}
then we augment \ADIMENSUMO{} by stating that $c_1$ and $c_2$ are related by {\it nonDisjoint}. By proceeding this way, non-disjointness is inherited both upwards and downwards, which enables a very compact formalization. In the above example, \SUMOClass{Beverage} and \SUMOClass{Water} are non-asserted {\it disjoint} (condition {\it a}) and, additionally, all the subclasses of \SUMOClass{Beverage} (\SUMOClass{Coffee}, \SUMOClass{Milk} and \SUMOClass{Tea}, among others) are non-asserted {\it disjoint} or not related with all the subclasses of \SUMOClass{Water} (condition {\it b}). Consequently, we augment \ADIMENSUMO{} as follows by the application of the CWA to {\it disjoint} assuming non-disjointness:
 \begin{equation} \label{formula:CWANonDisjointBeverageWater}
 nonDisjoint(Beverage,Water)
 \end{equation}
By formula (\ref{formula:CWANonDisjointBeverageWater}), \SUMOClass{Beverage} and \SUMOClass{Water} are directly asserted to be non-disjoint, as given by formula (\ref{formula:CWANonDisjointExplicitBeverageWater}). Further, all the pairs obtained from the super-classes of \SUMOClass{Beverage} and \SUMOClass{Water} respectively are asserted to be non-disjoint by upwards inheritance: for example, \SUMOClass{Beverage} and \SUMOClass{CompoundSubstance} (as given by formula (\ref{formula:CWANonDisjointExplicitBeverageCompoundSubstance})). Additionally, all the pairs obtained from the subclasses of \SUMOClass{Beverage} and \SUMOClass{Water} respectively are asserted to be non-disjoint by downwards inheritance: for example, \SUMOClass{Beverage} and \SUMOClass{Coffee} (as given by formula (\ref{formula:CWANonDisjointExplicitCoffeeWater})), \SUMOClass{Beverage} and \SUMOClass{Milk}, etc. Consequently, formulas (\ref{formula:CWANonDisjointExplicitBeverageCompoundSubstance}-\ref{formula:CWANonDisjointExplicitCoffeeWater}) (and many others) can be replaced with formulas (\ref{formula:nonDisjoint}-\ref{formula:CWANonDisjointBeverageWater}) while preserving logical equivalence. It is worth noting that condition {\it b} prevents the introduction of inconsistencies caused by the downwards inheritance of {\it non-disjoint}, since all the involved pairs of subclasses are restricted to be non-asserted {\it disjoint} or non related by {\it disjoint}.

In total, we have augmented \ADIMENSUMO{} by adding $29,643$ atomic formulas and $1$ general formula.

\section{Experimental Results} \label{section:experimentation}

\begin{table*}[ht!]
\caption{\label{table:newSolvedCQs} New solved CQs under the CWA}
\centering
\resizebox{2.05\columnwidth}{!}{
\begin{tabular}{l|rrrrrr|rrrrrr}
\toprule
\multicolumn{1}{c|}{\multirow{2}{*}{{\bf Competency}}} & \multicolumn{6}{c|}{\bf Passing} & \multicolumn{6}{c}{\bf Non-passing} \\ 
\cmidrule{2-13}
\multicolumn{1}{c|}{\multirow{2}{*}{{\bf Questions}}} & \multicolumn{3}{c}{\bf CWA-D} & \multicolumn{3}{c|}{\bf CWA-n-D} & \multicolumn{3}{c}{\bf CWA-D} & \multicolumn{3}{c}{\bf CWA-n-D} \\ 
\multicolumn{1}{c|}{\multirow{2}{*}{}} & \multicolumn{1}{c}{{\bf \#}} & \multicolumn{1}{c}{{\bf \%}} & \multicolumn{1}{c}{{\bf T}} & \multicolumn{1}{c}{{\bf \#}} & \multicolumn{1}{c}{{\bf \%}} & \multicolumn{1}{c|}{{\bf T}} & \multicolumn{1}{c}{{\bf \#}} & \multicolumn{1}{c}{{\bf \%}} & \multicolumn{1}{c}{{\bf T}} & \multicolumn{1}{c}{{\bf \#}} & \multicolumn{1}{c}{{\bf \%}} & \multicolumn{1}{c}{{\bf T}} \\
\midrule
noun \#1 (3,386) & 36 & 1.06~\% & 115.16 s. & 1,311 & 38.72~\% & 64.89 s. & 1,486 & 43.89~\% & 8.79 s. & 4 & 0.12~\% & 69.78 s. \\ 
noun \#2 (657) & 3 & 0.46~\% & 3.70 s. & 1 & 0.15~\% & 240.12 s. & 288 & 43.84~\% & 101.60 s. & 103 & 15.68~\% & 83.97 s. \\ 
verb \#1 (1,103) & 12 & 1.09~\% & 65.00 s. & 735 & 66.64~\% & 61.27 s. & 835 & 75.70~\% & 4.24 s. & 0 & 0.00~\% & 0.00 s. \\ 
verb \#2 (155) & 0 & 0.00~\% & 0.00 s. & 0 & 0.00~\% & 0.00 s. & 113 & 72.90~\% & 120.67 s. & 25 & 16.13~\% & 111.39 s. \\ 
antonym \#1 (29) & 1 & 3.45~\% & 3.74 s. & 0 & 0.00~\% & 0.00 s. & 1 & 3.45~\% & 10.25 s. & 1 & 3.45~\% & 40.39 s. \\ 
antonym \#2 (352) & 34 & 9.66~\% & 67.61 s. & 11 & 3.13~\% & 93.74 s. & 1 & 0.28~\% & 3.72 s. & 0 & 0.00~\% & 0.00 s. \\ 
antonym \#3 (1,357) & 326 & 24.02~\% & 107.35 s. & 626 & 46.13~\% & 110.31 s. & 0 & 0.00~\% & 0.00 s. & 0 & 0.00~\% & 0.00 s. \\ 
\midrule
{\bf Total (7,039)} & {\bf 412} & {\bf 5.85~\%} & {\bf 102.51 s.} & {\bf 2,684} & {\bf 38.13~\%} & {\bf 74.68 s.} & {\bf 2,724} & {\bf 38.70~\%} & {\bf 21.85 s.} & {\bf 133} & {\bf 1.89~\%} & {\bf 88.37 s.} \\ 
\bottomrule
\end{tabular}
}
\end{table*}

In this section, we present the experiments with the different versions of \ADIMENSUMO{} under the OWA and CWA as introduced in Section \ref{section:Completion}.

First of all, we have validated the completed versions of \ADIMENSUMO{} using white-box testing techniques \cite{AHL19}, and we have not found any inconsistency. Next, we have evaluated the efficiency  and competency of each FOL version of \ADIMENSUMO{}. For this purpose, we have used the framework for the evaluation of the competency of \SUMO{}-based ontologies introduced in \'Alvez {\it et al.} \cite{black19}. The interested reader can find a detailed analysis in \'Alvez {\it et al.} \cite{GWN19DetailedAnalysis}. This framework uses {\it competency questions} (CQs) \cite{GrF95} derived from several predefined question patterns (QPs) and three main knowledge resources: 1) the lexical database \WORDNET{} \cite{Fellbaum'98}, where lexical concepts encoded in {\it synonym sets} or {\it synsets} are semantically related by different types of semantic relations such as hyponymy, antonymy, meronymy, etc. 2) a FOL translation of \SUMO{} like \ADIMENSUMO{} and 3) the semantic mapping between \WORDNET{} and \SUMO{} \cite{Niles+Pease'03}. 
Specifically, our  benchmark is composed  of 14,324 commonsense CQs obtained from 4 QPs based on hyponymy ---2 QPs for nouns and 2 QPs for verbs--- and 3 QPs based on antonymy. Each CQ consists of two conjectures: the first is called the {\it truth-test}, which is expected to be entailed by the ontology and describes the CQ; the second is called {\it falsity-test}, which is obtained as the negation of the truth-test and is expected not to be entailed by the ontology. Next, we briefly describe our QPs and provide some examples of the resulting CQs. Given a hyponym pair of nouns or verbs, the semantics of the hyponym is subsumed by the semantics of the hyperonym, and our QPs simply state the same property in terms of \ADIMENSUMO{} depending on the mapping relation that is used for connecting the hyponym to \SUMO{}: two options, {\it instantiation}/{\it subsumption} or {\it equivalence}. For example, the synsets \synset{drinking\_water}{1}{n} and \synset{beverage}{1}{n} in Figure \ref{fig:introduction} are connected to \SUMOClass{Water} and \SUMOClass{Beverage} respectively, and the hyponym (i.e. \synset{drinking\_water}{1}{n}) is connected using {\it subsumption}. Thus, we apply the first QP based on {\it hyponymy} proposed in \cite{black19} and obtain a CQ consisting of the truth-test
\begin{equation}
\exists ?x \; ( \; instance(?x,Water) \wedge instance(?x,Beverage \; ) \label{CQ:WaterBeverage}
\end{equation}
and its negation. With respect to antonymy, the semantics of a pair of antonym synsets are incompatible, which is stated by our QPs in terms of \ADIMENSUMO{} depending on the mapping relations that are used for connecting the synsets to \SUMO{}. In this case, there are 3 options: in the two first ones the two synsets are connected using the same relation (either {\it instantiation}/{\it subsumption} or {\it equivalence}), and in the last one the synsets are connected using different relations (one is connected using {\it instantiation}/{\it subsumption} and the other one is connected using {\it equivalence}). For example, the adjectives \synset{liquescent}{1}{s} and \synset{frozen}{1}{a} are connected by {\it equivalence} to \SUMOClass{Melting} and \SUMOClass{Freezing} respectively. Thus, we apply the first QP based on {\it antonymy} proposed in \cite{black19} and obtain a CQ consisting of the truth-test
\begin{equation} \label{CQ:MeltingFreezing}
\begin{array}{l}
\forall ?x \forall ?y \; ( \; ( \; instance(?x,Melting) \; \wedge \\
\hspace{44pt} instance(?y,Freezing) \; ) \; \to \; \neg ?x = ?y \; ) 
\end{array}
\end{equation}
and its negation.

Given a benchmark, two dual tests are performed for each CQ using FOL ATPs: the first test is to check whether, as expected, the truth-test is entailed by the ontology; the second one is to check if the falsity-test is entailed. If ATPs find a proof for either the truth- or the falsity-test, then the CQ is classified as {\it solved} (or {\it resolved}). In particular, the CQ is {\it passing}/{\it non-passing} if ATPs find a proof for the truth-test/falsity-test. Otherwise (that is, if no proof is found), the CQ is classified as {\it unresolved} or {\it unknown}.\footnote{Given a consistent ontology, ATPs cannot find a proof for both the truth- and the falsity-test.} For example, the CQ described by ``{\it Drinking water is a beverage}'', which consists of truth-test (\ref{CQ:WaterBeverage}) and its negation, is classified as {\it unknown} by using the original version of \ADIMENSUMO{}, as {\it non-passing} by using \ADIMENSUMO{} augmented by the application of the CWA to {\it subclass} and {\it disjoint} assuming disjointness, and as {\it passing} by using \ADIMENSUMO{} augmented by the application of the CWA to {\it subclass} and {\it disjoint} assuming non-disjointness.

Our experimentation has been performed by using Vampire v4.2.2 ---which is the {\it CADE ATP System Competition} (CASC) FOF\footnote{First-Order Form non-propositional theorems (axioms with a provable conjecture).} division winner in 2017 \cite{PSS02,SuS06} and the latest available stable release\footnote{\url{https://vprover.github.io/}} of Vampire at the time of our experimentation--- in a Intel\textregistered~Xeon\textregistered~CPU E5-2640v3@2.60GHz with 2GB of RAM memory per processor. For each test, we have set an execution-time limit of 300 seconds and a memory limit of 2GB.\footnote{Parameters: \tt{--proof tptp --output\_axiom\_names on --mode casc -t 300 -m 2048}} Totally, the experimentation has required almost 300 days/processor of computation effort: 3 ontologies, 14,324 CQs, 2 tests per CQ and 300 seconds per test. All the required knowledge resources ---the original ontology \ADIMENSUMO{} and its versions under the CWA, the set of CQs and conjectures, the mapping between \SUMO{} and \WORDNET{} v3.0, \WORDNET{} v3.0 relation pairs--- and the resulting execution reports are available at \url{https://adimen.si.ehu.es/web/AdimenSUMO}.

We summarize our experimental results in Table \ref{table:summary}, where CQs are organized by QP. In the first column (Competency Questions column), we provide the QP type and the number of CQs (between brackets). In the next 9 columns, we provide the number (columns \#), percentage (columns \%) and average runtime (columns T) of CQs that are solved by using each version of \ADIMENSUMO{}: the original version of \ADIMENSUMO{} (OWA, 3 columns), \ADIMENSUMO{} augmented by applying the CWA to {\it subclass} and {\it disjoint} assuming disjointness (CWA-D, 3 columns), and \ADIMENSUMO{} augmented by applying the CWA to {\it subclass} and {\it disjoint} assuming non-disjointness (CWA-n-D, 3 columns).


From our results, it is easy to see that the CWA surpasses the OWA in our benchmark in terms of competency: the two augmented versions of \ADIMENSUMO{} outperform the original version in terms of solved CQs ($9,781$ and $9,498$ against $7,285$ solved CQs) and that the total number of solved CQs increases more than $50$~\% ($10,970$ against $7,285$ solved CQs). Further, for each QP the number of solved CQs also increases up to $136~\%$ ($1,502$ against $637$ solved CQs from {\it verb \#1}). However, in the case of the CQs that result from {\it antonym \#1}, the original version of \ADIMENSUMO{} outperforms the augmented ones ($36$ against $29$ and $27$ solved CQs). This is because when using the augmented versions of \ADIMENSUMO{} the ATP runs out of resources (mainly time) at trying to solve some of the CQs that were already solved by using the original version of the ontology. In total, $359$ CQs ($4.93~\%$) that are solved by using the original version of \ADIMENSUMO{} remain unresolved when trying one of the augmented versions of \ADIMENSUMO{} in our experimentation. However, for each QP many new CQs are solved only when using one of the augmented versions of \ADIMENSUMO{}. Moreover,  even improving the competency, the augmented versions of \ADIMENSUMO{} also outperforms the original one in terms of efficiency (55.12 s. and 58.75 s. against 61.30 s.), mainly because of the efficiency improvement at solving the CQs obtained from {\it verb \#1} and {\it antonym \#3}. This implies that the new added knowledge has not a deep negative impact in the efficiency of the augmented ontologies. Further, we have checked that the newly added axioms sometimes serve as shortcuts in the proof of problems that were already solved using the original version of \ADIMENSUMO{}.

Additionally, in Table \ref{table:newSolvedCQs} we provide some figures about the CQs that remain unresolved when using the original version of \ADIMENSUMO{}. In the first column (Competency Questions column), we provide the QP from which CQs have been obtained and the number of CQs that remain unresolved when using the original version of \ADIMENSUMO{} (between brackets). 
The last 12 columns are organized into groups of 3 columns. In each group, we provide the number (\# columns), percentage of CQs (\% columns) and average runtime (T columns) that are respectively classified as {\it passing}/{\it non-passing} by using each augmented version of \ADIMENSUMO{}: \ADIMENSUMO{} augmented by applying the CWA to {\it subclass} and {\it disjoint} assuming disjointness (CWA-D), and \ADIMENSUMO{} augmented by applying the CWA to {\it subclass} and {\it disjoint} assuming non-disjointness (CWA-n-D).

According to the reported results, the classification of the newly solved CQs strongly depends on the given assumption that we adopt in order to apply the CWA: $2,684$ CQs ($38.13~\%$) are classified as {\it passing} if assuming non-disjointness, while $2,724$ CQs ($38.70~\%$) are classified as {\it non-passing} if assuming disjointness. Among them, $732$ CQs are solved only when applying the CWA to {\it subclass} and {\it disjoint} by assuming disjointness. On the contrary, $527$ CQs are solved only when applying the CWA to {\it subclass} and {\it disjoint} by assuming non-disjointness. Further, the chosen assumption also influences the kind of CQs that are most frequently solved: the largest amount of CQs obtained from hyponymy are solved when assuming disjointness ($2,773$ against $2,179$ solved CQs obtained from {\it noun \#1}, {\it noun \#2}, {\it verb \#1} and {\it verb \#2}), although the largest number of CQs obtained from antonymy-based QPs are solved when assuming non-disjointness ($638$ against $363$ solved CQs obtained from {\it antonym \#1}, {\it antonym \#2} and {\it antonym \#3}). Regarding average runtimes, it seems that assuming disjointness at applying the CWA to {\it disjoint} yields to a more efficient augmented version of \ADIMENSUMO{}.

\section{Discussion} \label{section:discussion}

In this section, we discuss the experimental results reported in the above section.

The experimental results reported in Section \ref{section:experimentation} can be further improved. First, we think that the ATP runs out of resources (especially time) when trying to prove conjectures that are entailed by some of the augmented versions of \ADIMENSUMO{}. Actually, we have experimentally checked that the ATP runs out of resources using the augmented versions of \ADIMENSUMO{} when trying to solve $359$ CQs that are solved by the original version of \ADIMENSUMO{} (less than $5~\%$). Thus, it is very likely that there are more solvable CQs. Second, we also think that our results are penalized by the poor mapping of adjective synsets as pointed out by \cite{GWN19DetailedAnalysis}, since the worst results reported in Table \ref{table:summary} correspond to the CQs obtained from the QPs based on pairs of {\it antonym} adjectives. Third, we have manually inspected some cases and   detected that some knowledge is still under-specified, despite of the application of the CWA. For example, it is not possible to infer from the augmented versions of \ADIMENSUMO{} whether \SUMOClass{Animal} and \SUMOClass{LinguisticExpression} are {\it disjoint} or not. 
Another example of missing knowledge is the axiomatiation of many attributes. We have discovered this problem by analysing the most frequent concepts involved in the {\it unknown} CQs e.g.   
\SUMOClass{SubjectiveAssessmentAttribute}. 

Regarding non-asserted {\it subclass} pairs, our assumption is that those pairs are not related by {\it subclass}, since otherwise most of the involved classes would become equal by anti-symmetry as discussed in Section \ref{section:Completion}. Interestingly, the impact of augmenting \ADIMENSUMO{} by applying the CWA only to {\it subclass} as described in Subsection \ref{subsection:CompletingSubclass} is really small. On the contrary, the impact of the application of the CWA to {\it subclass} in combination with the application of the CWA to {\it disjoint} is much higher. This is especially the case when applying the CWA by assuming disjointness, mainly because the DCA is also applied in the proposal described in Subsection \ref{subsection:CompletingSubclass}.

With respect to non-asserted {\it disjoint} pairs, we have assumed that those pairs are either disjoint (in Subsection \ref{subsection:CompletingDisjoint}) or non-disjoint (in Subsection \ref{subsection:CompletingNonDisjoint}). As described in Section \ref{section:experimentation}, the classification of most of the newly solved CQs depends strongly on the chosen assumption: if assuming disjointness, $2,724$ newly solved CQs are classified as {\it non-passing}, while $3,136$ newly solved CQs are classified as {\it passing} when assuming non-disjointness (see Table \ref{table:newSolvedCQs}). An example of this is the CQ described in the introduction: ``{\it Drinking water is a beverage}'', see truth-test (\ref{CQ:WaterBeverage}). This fact confirms the lack of structural knowledge about classes in \ADIMENSUMO{}.

We have, therefore, proved that structural knowledge is missing and it will be necessary to augment the ontology. We foresee that the classification of the newly solved CQs can guide the application of the CWA to {\it disjoint} using \WORDNET{} as a reliable knowledge source. Let us explain this proposal with two examples. On the one hand, by assuming that \SUMOClass{Melting} and \SUMOClass{Freezing} are disjoint, the CQ described by ``{\it the adjectives liquescent and frozen are antonym}'', consisting of truth-test (\ref{CQ:MeltingFreezing}) and its negation, is {\it passing}, but if non-disjointness is assumed, it is {\it non-passing}. So, using the knowledge in the ontology and the combination of both augmented versions we can conclude that \SUMOClass{Melting} and \SUMOClass{Freezing} must be {\it disjoint}. On the other hand, the CQ described by ``{\it Drinking water is a beverage}'' is  {\it non-passing} when assuming that \SUMOClass{Beverage} and \SUMOClass{CompoundSubstance} are disjoint but {\it passing} when assuming non-disjointness. Thus, these classes should be {\it non-disjoint}. In sum, the combination of assuming disjointness in cases such as \SUMOClass{Melting} and \SUMOClass{Freezing} 
and non-disjointness in cases such as \SUMOClass{Beverage} and \SUMOClass{CompoundSubstance} seems to be appropriate to obtain the correct disjoint/non-disjoint axioms if using \WORDNET{} as a reliable knowledge source. 
In any case, the above described solutions for {\it disjoint} require the addition of many new axioms. We have proved that the source \ADIMENSUMO{} and its completed versions are comparable in terms of efficiency according to the experimentation reported in Section \ref{section:experimentation}, even though we have added many axioms. However, if we follow  suitable structural design criteria, it is not necessary  to include additional axioms in a FOL ontology for the application of the CWA to {\it disjoint}.

One of these possible criteria is linked to the  the application of the CWA to {\it subclass}: Organizing the  knowledge around the notion of classes will implicitly provide a solution for {\it disjoint}. If we define two classes as disjoint iff they do not share any common subclass, the completion of {\it subclass} itself (see Subsection \ref{subsection:CompletingSubclass}) enables deciding whether two classes are disjoint or not. However, at this time, this is not possible because the notion of disjointness in \SUMO{} inappropriately states that two classes are disjoint iff they do not share any common instance \begin{equation} \label{formula:SUMODisjoint}
\begin{array}{l}
\forall ?x_1 \forall ?x_2 \forall ?y \; ( \; disjoint(?x_1,?x_2) \; \to \\
\hspace{96pt }\neg \; ( \; instance(?y,?x_1) \; \wedge \\
\hspace{111pt} instance(?y,?x_2) \; ) \; ) \\
\end{array}
\end{equation}
and not subclasses as we propose:
\begin{equation} \label{formula:SUMODisjoint}
\begin{array}{l}
\forall ?x_1 \forall ?x_2 \forall ?y \; ( \; disjoint(?x_1,?x_2) \; \to \\
\hspace{96pt }\neg \; ( \; subclass(?y,?x_1) \; \wedge \\
\hspace{111pt} subclass(?y,?x_2) \; ) \; ) \\
\end{array}
\end{equation}

Further, the inappropriate use of {\it instance} is extended to most parts of \SUMO{} and makes it possible to infer that many pairs of classes that do not share any common subclass do have common instances. This fact  makes it really difficult to correct the axiomatization of {\it disjoint} in \SUMO{} without reconstructing all the knowledge from almost scratch and, consequently, it prevents the easy application of CWA  to {\it disjoint} in \SUMO{}. 

In sum, we consider that the best choice for the practical application of the CWA  in a FOL ontology is to follow the suitable design criteria such as the one explained above. 

\section{Conclusions and Future Work} \label{section:Conclusions}

To the best of our knowledge, up to now the research and evaluation of \SUMO{}-based FOL ontologies have been developed exclusively under the {\it Open World Assumption}. 
This paper reports on the first investigation on the application of the {\it Closed World Assumption} to \SUMO{}-based FOL ontologies. Concretely, we have applied the Careful CWA introduced by \cite{gelfond1986negation} to the {\it subclass} and {\it disjoint} relations of a FOL version of \SUMO{}. We have checked two CWA formulations for {\it disjoint}: i) by assuming disjointness and ii) by assuming non-disjointness. We have tested these two formulations on a very large benchmark of 14,324 commonsense competency questions extracted from \WORDNET{} and its mapping to \SUMO{}. Summing up, although the size of the ontologies has been increased, the resulting ontologies are far more competent and keep their efficiency. Regardless of the CWA strategy applied to {\it disjoint}, our research empirically demonstrates that the competency of the ontology can improve more than $50$~\% when reasoning under the CWA. As a side effect, we have also discovered that the {\it missing} structural knowledge in \SUMO{} is nowadays one of the main obstacles for its practical application in automated commonsense reasoning tasks. In fact, almost $30$~\% of the structural knowledge about classes is missing in \ADIMENSUMO{}. Further, our proposal can help ontologists to complete the missing knowledge, for example by using \WORDNET{}. Thus, the practical utility of our proposal in tasks that require commonsense reasoning is clear.

Although the approach assuming disjointness obtains the best results, a combination of both approaches should be further investigated. For example, using \WORDNET{} as reliable source of knowledge, a possible approach is to weigh each new pair according to the number of solutions in which it is used and its kind ({\it passing}/{\it non-passing}), and then proceed to choose the most relevant ones while keeping consistency. Similar approaches could be taken into account by considering other sources of knowledge. Additionally, as discussed, suitable design criteria can facilitate the application of the CWA to FOL ontologies. Future work will focus on implementing the proposed strategies. It will also involve experimenting with other knowledge representation strategies such as the {\it Unique Name Assumption} (UNA) \cite{LMP08} and testing augmented versions of the ontology with other datasets such us the ones created in the Webchild project \cite{tandon2017webchild} and ConceptNet \cite{speer2017conceptnet}.

\ack We would like to thank the reviewers for their comments, which helped improve this paper considerably.

This work has been partially funded by the the project DeepReading (RTI2018-096846-B-C21) supported by the Ministry of Science, Innovation and Universities of the Spanish Government, and GRAMM (TIN2017-86727-C2-2-R) supported  by the Ministry of  Economy, Industry and Competitiveness of the Spanish Government, the Basque Project LoRea (GIU18/182), Ixa Group-consolidated group type A by the Basque Government (IT1343-19) and BigKnowledge -- {\it Ayudas Fundaci\'on BBVA a Equipos de Investigaci\'on Cient\'ifica 2018}.

\bibliography{references}

\begin{thebibliography}{10}

\bibitem{AGR18}
J.~{\'A}lvez, I.~Gonzalez-Dios, and G.~Rigau, `Cross-checking {\WORDNET{}} and
  {\SUMO{}} using meronymy', in {\em Proc. of the 11$^{th}$ {Int. Conf. on
  Language Resources and Evaluation (LREC 2018)}}, pp. 4570--4577, (2018).

\bibitem{GWN19DetailedAnalysis}
J.~{\'A}lvez, I.~Gonzalez-Dios, and G.~Rigau, `Commonsense reasoning using
  {\WORDNET{}} and {\SUMO{}}: a detailed analysis', in {\em Proc. of the
  10$^{th}$ Global WordNet Conference (GWC 2019)}, pp. 197--205, (2019).

\bibitem{AHL19}
J.~{\'{A}}lvez, M.~Hermo, P.~Lucio, and G.~Rigau, `Automatic white-box testing
  of first-order logic ontologies', {\em Journal of Logic and Computation},
  {\bf 29}(5),  723--751, (2 2019).

\bibitem{ALR12}
J.~{\'A}lvez, P.~Lucio, and G.~Rigau, `{\ADIMENSUMO{}}: Reengineering an
  ontology for first-order reasoning', {\em Int. J. Semantic Web Inf. Syst.},
  {\bf 8}(4),  80--116, (2012).

\bibitem{ALR17}
J.~{\'A}lvez, P.~Lucio, and G.~Rigau, `Black-box testing of first-order logic
  ontologies using {\WORDNET{}}', {\em CoRR}, {\bf abs/1705.10217}, (2017).

\bibitem{black19}
J.~\'{A}lvez, P.~Lucio, and G.~Rigau, `A framework for the evaluation of
  {\SUMO{}}-based ontologies using {\WORDNET{}}', {\em IEEE Access}, {\bf 7},
  36075--36093, (2019).

\bibitem{AnH08}
G.~Antoniou and F.~Harmelen, {\em A Semantic Web Primer (2$^{nd}$ edition)},
  MIT Press, 2008.

\bibitem{Cha99}
B.~Chandrasekaran, John~R. Josephson, and V.~Richard Benjamins, `What are
  ontologies, and why do we need them?', {\em IEEE Intelligent Systems}, {\bf
  14}(1),  20--26, (1999).

\bibitem{Cla78}
K.~L. Clark, {\em Negation As Failure},  293--322, Springer, Boston, MA, 1978.

\bibitem{Aquin11}
M.~d'Aquin and N.~F. Noy, `Where to publish and find ontologies? {A} survey of
  ontology libraries', {\em Web Semantics: Science, Services and Agents on the
  World Wide Web}, {\bf 11},  96--111, (2012).

\bibitem{drummond2006open}
N.~Drummond and R.~Shearer, `The open world assumption', in {\em eSI Workshop:
  The Closed World of Databases meets the Open World of the Semantic Web},
  volume~15, (2006).

\bibitem{EIL08}
T.~Eiter, G.~Ianni, T.~Lukasiewicz, R.~Schindlauer, and H.~Tompits, `Combining
  answer set programming with description logics for the semantic web', {\em
  Artificial Intelligence}, {\bf 172}(12),  1495 -- 1539, (2008).

\bibitem{Fellbaum'98}
{\em {WordNet:} {A}n Electronic Lexical Database}, ed., C.~Fellbaum, MIT Press,
  1998.

\bibitem{Gangemi+'02}
A.~Gangemi, N.~Guarino, C.~Masolo, A.~Oltramari, and L.~Schneider, `Sweetening
  ontologies with {DOLCE}', in {\em {Knowledge Engin. and Knowledge Manag.:
  Ontologies and the Semantic Web}}, ed., A.~{G{\'o}mez-P{\'e}rez et al.}, LNCS
  2473,  166--181, Springer, (2002).

\bibitem{gelfond1986negation}
M.~Gelfond and H.~Przymusinska, `Negation as failure: Careful closure
  procedure', {\em Artificial Intelligence}, {\bf 30}(3),  273--287, (1986).

\bibitem{Richard+'92}
M.~R. Genesereth, R.~E. Fikes, D.~Brobow, R.~Brachman, T.~Gruber, P.~Hayes,
  R.~Letsinger, V.~Lifschitz, R.~Macgregor, J.~Mccarthy, P.~Norvig, R.~Patil,
  and L.~Schubert, `{Knowledge Interchange Format} version 3.0 reference
  manual', Technical Report Logic-92-1, Stanford University, Computer Science
  Department, Logic Group, (1992).

\bibitem{Grimm2007knowledge}
S.~Grimm, P.~Hitzler, and A.~Abecker, `Knowledge representation and
  ontologies', {\em Semantic Web Services: Concepts, Technologies, and
  Applications},  51--105, (2007).

\bibitem{GrF95}
M.~Gr{\"u}ninger and M.~S. Fox, `Methodology for the design and evaluation of
  ontologies', in {\em Proc. of {the Workshop on Basic Ontological Issues in
  Knowledge Sharing (IJCAI 1995)}}, (1995).

\bibitem{HoV11}
K.~Hoder and A.~Voronkov, `Sine qua non for large theory reasoning', in {\em
  Proc. of the 23$^{rd}$ {Int. Conf. on Automated Deduction (CADE-23)}}, eds.,
  N.~Bj{\o}rner and V.~Sofronie-Stokkermans, volume 6803 of {\em Lecture Notes
  in Computer Science},  299--314, Springer Berlin Heidelberg, (2011).

\bibitem{HoV06}
I.~Horrocks and A.~Voronkov, `Reasoning support for expressive ontology
  languages using a theorem prover', in {\em Foundations of Information and
  Knowledge Systems}, ed., {J. Dix et al.}, LNCS 3861,  201--218, Springer,
  (2006).

\bibitem{KoV13}
L.~Kov\'{a}cs and A.~Voronkov, `First-order theorem proving and {Vampire}', in
  {\em {Computer Aided Verification}}, eds., N.~Sharygina and H.~Veith, LNCS
  8044,  1--35, Springer, (2013).

\bibitem{KSH11}
A.-A. Krisnadhi, K.~Sengupta, and P.~Hitzler, `Local closed world semantics:
  Keep it simple, stupid!', in {\em {Proc. of the 24$^{th}$ Int. Workshop on
  Description Logics (DL 2011)}}, eds., R.~Rosati, S.~Rudolph, and
  M.~Zakharyaschev, volume 745 of {\em {CEUR} Workshop Proceedings}.
  CEUR-WS.org, (2011).

\bibitem{LMP08}
V.~Lifschitz, L.~Morgenstern, and D.~Plaisted, `Knowledge representation and
  classical logic', in {\em Handbook of Knowledge Representation}, eds., F.~van
  Harmelen, V.~Lifschitz, and B.~Porter,  3--88, Elsevier, (2008).

\bibitem{Matuszek+'06}
C.~Matuszek, J.~Cabral, M.~J. Witbrock, and J.~DeOliveira, `An introduction to
  the syntax and content of {Cyc}', in {\em Proc. of the {Spring Symposium:
  Formalizing and Compiling Background Knowledge and Its Appl. to Knowledge
  Repr. and Question Answering}}, ed., C.~Baral, pp. 44--49. AAAI Press,
  (2006).

\bibitem{MoR08}
B.~Motik and R.~Rosati, `Reconciling description logics and rules', {\em J.
  ACM}, {\bf 57}(5),  30:1--30:62, (June 2008).

\bibitem{Niles+Pease'01}
I.~Niles and A.~Pease, `Towards a standard upper ontology', in {\em Proc. of
  the 2$^{nd}$ {Int. Conf. on Formal Ontology in Information Systems (FOIS
  2001)}}, ed., {Guarino N. et al.}, pp. 2--9. ACM, (2001).

\bibitem{Niles+Pease'03}
I.~Niles and A.~Pease, `Linking lexicons and ontologies: Mapping {WordNet} to
  the {Suggested Upper Merged Ontology}', in {\em Proc. of the {IEEE Int. Conf.
  on Inf. and Knowledge Engin. (IKE 2003)}}, ed., H.~R. Arabnia, volume~2, pp.
  412--416. CSREA Press, (2003).

\bibitem{noy2001ontology}
N.~F. Noy and D.~L. McGuinness, `Ontology development 101: A guide to creating
  your first ontology', Technical Report KSL-01-05 and SMI-2001-0880, Stanford
  Knowledge Systems Laboratory and Stanford Medical Informatics, (2001).

\bibitem{Pea09}
A.~Pease, `{Standard Upper Ontology Knowledge Interchange Format}'.
\newblock Retrieved June 18, 2009, from
  \url{http://sigmakee.cvs.sourceforge.net/sigmakee/sigma/suo-kif.pdf}, 2009.

\bibitem{PeS07}
A.~Pease and G.~Sutcliffe, `First-order reasoning on a large ontology', in {\em
  Proc. of the {Workshop on Empirically Successful Automated Reasoning in Large
  Theories (CADE-21)}}, ed., {Sutcliffe G. et al.}, CEUR Workshop Proceedings
  257, CEUR-WS.org, (2007).

\bibitem{pease2010large}
A.~Pease, G.~Sutcliffe, N.~Siegel, and S.~Trac, `{Large theory reasoning with
  SUMO at CASC}', {\em AI Communications}, {\bf 23}(2-3),  137--144, (2010).

\bibitem{PSS02}
F.J. Pelletier, G.~Sutcliffe, and C.B. Suttner, `{The Development of CASC}',
  {\em AI Communications}, {\bf 15}(2-3),  79--90, (2002).

\bibitem{RRG05}
D.~Ramachandran, R.~P. Reagan, and K.~Goolsbey, `First-orderized {ResearchCyc}:
  Expressivity and efficiency in a common-sense ontology', in {\em Papers from
  the {Workshop on Contexts and Ontologies: Theory, Practice and Applications
  (AAAI 2005)}}, ed., {P. Shvaiko et al.}, pp. 33--40. AAAI Press, (2005).

\bibitem{Rei78}
R.~Reiter, {\em On Closed World Data Bases},  55--76, Springer, Boston, MA,
  1978.

\bibitem{reiter1980logic}
R.~Reiter, `A logic for default reasoning', {\em Artificial intelligence}, {\bf
  13}(1-2),  81--132, (1980).

\bibitem{Sch02}
S.~Schulz, `{E - A} brainiac theorem prover', {\em AI Communications}, {\bf
  15}(2-3),  111--126, (2002).

\bibitem{Siebert19}
S.~Siebert, C.~Schon, and F.~Stolzenburg, `Commonsense reasoning using theorem
  proving and machine learning', in {\em Machine Learning and Knowledge
  Extraction}, eds., A.~Holzinger, P.~Kieseberg, A~M. Tjoa, and E.~Weippl, pp.
  395--413, Cham, (2019). Springer International Publishing.

\bibitem{speer2017conceptnet}
R.~Speer, J.~Chin, and C.~Havasi, `{ConceptNet} 5.5: An open multilingual graph
  of general knowledge', in {\em Thirty-First AAAI Conference on Artificial
  Intelligence}, (2017).

\bibitem{StS09}
S.~Staab and R.~Studer, {\em Handbook on Ontologies}, Springer Publishing
  Company, Incorporated, 2$^{nd}$ edn., 2009.

\bibitem{Sut09}
G.~Sutcliffe, `The {TPTP} problem library and associated infrastructure', {\em
  J. Automated Reasoning}, {\bf 43}(4),  337--362, (2009).

\bibitem{SuS06}
G.~Sutcliffe and C.~Suttner, `{The State of CASC}', {\em AI Communications},
  {\bf 19}(1),  35--48, (2006).

\bibitem{tandon2017webchild}
N.~Tandon, G.~De Melo, and G.~Weikum, `{WebChild} 2.0: Fine-grained commonsense
  knowledge distillation', in {\em Proc. of ACL 2017, System Demonstrations},
  pp. 115--120, (2017).

\bibitem{vossen2016newsreader}
P.~Vossen, R.~Agerri, I.~Aldabe, A.~Cybulska, M.~van Erp, A.~Fokkens,
  E.~Laparra, A.~Minard, A.~Palmero Aprosio, and G.~Rigau, `Newsreader: Using
  knowledge resources in a cross-lingual reading machine to generate more
  knowledge from massive streams of news', {\em Knowledge-Based Systems}, {\bf
  110},  60--85, (2016).

\bibitem{W3C12}
{W3C OWL Working Group}.
\newblock {OWL 2 Web Ontology Language Document Overview (Second Edition) - W3C
  Recommendation 11 December 2012}, 2012.

\end{thebibliography}
\end{document}